\documentclass[conference]{IEEEtran}
\usepackage{graphicx}
\usepackage{amsmath}
\usepackage{amssymb}
\usepackage{booktabs}
\usepackage{hyperref}
\usepackage{algorithm}
\usepackage{algorithmic}
\usepackage[table]{xcolor}
\usepackage{multirow}
\usepackage{float} 

\begin{document}

\title{MotionSwap: An Efficient Framework for High Fidelity Face Swapping with Attention-Enhanced Generator}

\author{
\IEEEauthorblockN{Om Patil$^{1}$, Jinesh Modi$^{1}$, Suryabha Mukhopadhyay$^{1}$, Meghaditya Giri$^{1}$, Chhavi Malhotra$^{1}$}
\IEEEauthorblockA{$^{1}$BITS Pilani, Hyderabad Campus\\
\{f20220046, f20220084, f20220166, f20220155, f20221386\}@hyderabad.bits-pilani.ac.in}
}

\maketitle

\begin{abstract}
Face swapping technology has gained significant attention in both academic research and commercial applications. This paper presents our implementation and enhancement of SimSwap, an efficient framework for high fidelity face swapping. We introduce several improvements to the original model, including self and cross-attention mechanisms in the generator architecture, dynamic loss weighting, and cosine annealing learning rate scheduling. Our enhanced model demonstrates significant improvements in identity preservation and attribute consistency compared to the original implementation. We provide a comprehensive analysis of the training process, showing how the model performance evolves through 400,000 iterations. The experimental results validate the effectiveness of our approach, while we also identify promising directions for future work, including the integration of StyleGAN3, improved lip synchronization, and 3D facial modeling.
\end{abstract}

\section{Introduction}
Face swapping involves transferring the identity of a source face to a target face while preserving the original expressions, poses, and lighting conditions of the target. This technology has numerous applications, ranging from entertainment and film production to privacy protection and virtual try-on systems. However, achieving high-fidelity face swapping that maintains both identity and attribute consistency remains challenging.

SimSwap, introduced by Chen et al. \cite{chen2020simswap}, represents a milestone in face swapping technology, offering an efficient framework that addresses many limitations of previous approaches. In this paper, we present our implementation and enhancement of the SimSwap framework, focusing on improving the quality and efficiency of face swapping through architectural improvements and optimization strategies.

Our main contributions include:
\begin{itemize}
    \item Integration of self and cross-attention mechanisms into the generator architecture to enhance feature processing and identity transfer
    \item Implementation of dynamic loss weighting to balance identity preservation and attribute consistency
    \item Adoption of cosine annealing learning rate scheduling to improve training stability and convergence
    \item Comprehensive analysis of the training process over 400,000 iterations
    \item Identification of promising directions for future improvement
\end{itemize}

The remainder of this paper is organized as follows: Section II reviews related work in face swapping and deepfake generation. Section III details our methodology, including problem formulation, model architecture, and loss functions. Section IV presents the implementation details and experimental setup. Section V discusses the results of our experiments. Section VI analyzes the strengths and limitations of our approach and outlines directions for future work. Finally, Section VII concludes the paper.

\section{Background and Related Work}
\subsection{Deep Learning for Deepfakes}
The creation of realistic face manipulations, commonly known as "deepfakes," has advanced significantly with the development of deep learning techniques. Nguyen et al. \cite{nguyen2019deep} provide a comprehensive survey of deep learning methods for deepfake creation and detection. These techniques typically leverage generative models to synthesize realistic facial imagery, enabling various applications from entertainment to privacy protection.

The rapid advancement of deepfake technology has also sparked interest in detection methods. Li et al. \cite{li2020learn} proposed a facial knowledge distillation approach for deepfake detection, while Afchar et al. \cite{afchar2018mesonet} developed MesoNet, a compact network specifically designed for video forgery detection. Nguyen et al. \cite{nguyen2019capsule} explored the use of capsule networks for detecting forged images and videos.

\subsection{Face Swapping Approaches}
Several approaches have been proposed for face swapping, each with its unique strengths and limitations:

\subsubsection{StyleGAN-based Methods}
Karras et al. \cite{karras2019style} introduced StyleGAN, a style-based generator architecture that allows for fine control over facial attributes. This approach enables high-quality face synthesis and manipulation, including identity transfer. The style-based approach separates high-level attributes from stochastic variations, facilitating more precise control over the generated images.

\subsubsection{Face2Face}
Thies et al. \cite{thies2016face2face} developed Face2Face, a method for real-time face capture and reenactment of RGB videos. Unlike pure face swapping, Face2Face focuses on expression transfer, preserving the identity of the target face while transferring expressions from a source face. This approach uses a 3D morphable face model to capture facial expressions and transfer them between individuals.

\subsubsection{Neural Voice Puppetry}
Thies et al. \cite{thies2020neural} extended their work to audio-driven facial reenactment with Neural Voice Puppetry. This method connects speech to facial expressions, enabling the generation of realistic talking head videos from audio input. While not directly focused on face swapping, this technology shares many technical foundations with face swapping approaches.

\subsubsection{First Order Motion Model}
Siarohin et al. \cite{siarohin2019first} presented the First Order Motion Model for image animation, which can be applied to face animation. This approach uses keypoint detection and local affine transformations to transfer motion between objects, allowing for realistic animation of still images using a driving video.

\subsection{MotionSwap Framework}
SimSwap \cite{chen2020simswap} addresses limitations in previous face swapping approaches by designing an efficient framework that preserves both identity similarity and attribute consistency. It introduces an identity injection module that transfers identity information while maintaining the original attributes of the target face. The framework employs a weakly supervised learning approach without requiring paired training data, making it more practical for real-world applications.

Unlike previous methods that often sacrifice attribute consistency for identity preservation, MotionSwap achieves a better balance between these two aspects. It uses a novel identity injection module and carefully designed loss functions to maintain both the identity of the source face and the attributes of the target face. The framework also employs adversarial training to enhance the realism of the generated faces.

\section{Methodology}
\subsection{Problem Formulation}
Given a source image $I_s$ containing the desired identity and a target image $I_t$ with the desired attributes (expression, pose, lighting), our goal is to generate a swapped face $I_{s\rightarrow t}$ that preserves the identity from $I_s$ while maintaining the attributes from $I_t$. This can be formulated as finding a function $F$ such that:

\begin{equation}
I_{s\rightarrow t} = F(I_s, I_t)
\end{equation}

where $I_{s\rightarrow t}$ should satisfy:
\begin{itemize}
    \item Identity similarity: $\text{id}(I_{s\rightarrow t}) \approx \text{id}(I_s)$
    \item Attribute consistency: $\text{attr}(I_{s\rightarrow t}) \approx \text{attr}(I_t)$
\end{itemize}

Here, $\text{id}(\cdot)$ represents the identity information and $\text{attr}(\cdot)$ represents the attribute information of a face image.

\subsection{Model Architecture}
\subsubsection{Generator Architecture}
Our generator follows an encoder-decoder structure with an identity injection module. The encoder extracts feature representations from both source and target images, while the decoder reconstructs the swapped face image. The identity injection module transfers identity information from the source to the target features.

The enhanced generator can be formulated as:
\begin{equation}
I_{s\rightarrow t} = G(I_s, I_t, \theta_G)
\end{equation}

where $\theta_G$ represents the generator parameters.

The generator architecture consists of the following components:
\begin{itemize}
    \item Encoder: A series of convolutional layers that extract feature maps from input images
    \item Identity Injection Module: Transfers identity information from source to target features
    \item Attention Modules: Self and cross-attention layers to enhance feature processing
    \item Residual Blocks: Multiple residual blocks for feature transformation
    \item Decoder: Upsampling and convolutional layers that reconstruct the output image
\end{itemize}

\begin{figure}[t]
\centering
\includegraphics[width=0.9\linewidth]{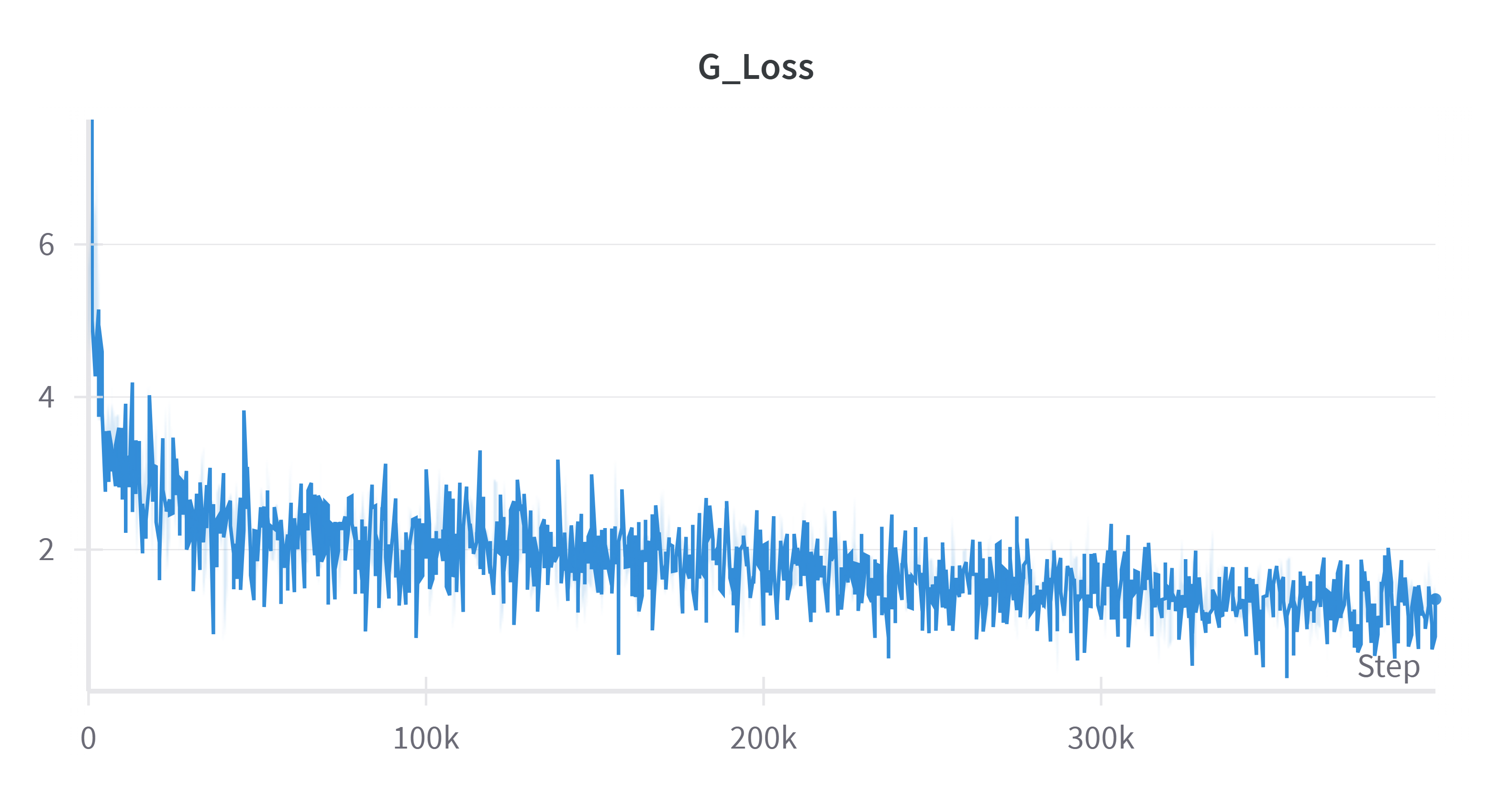} 
\caption{Loss curves during training showing the progression of generator losses over 400,000 iterations. The steady decline indicates effective model optimization and convergence.}
\label{fig:loss_curves}
\end{figure}

\subsubsection{Identity Extractor}
We utilize ArcFace (ResNet-100) as our identity feature extractor, which produces a 512-dimensional identity embedding:
\begin{equation}
f_{\text{id}}(I) = E_{\text{id}}(I)
\end{equation}

The identity embedding is used both for the identity injection module and for computing the identity loss during training.

\subsubsection{Discriminator}
Our discriminator follows a multi-scale architecture with patch-based discrimination. It consists of several convolutional layers followed by leaky ReLU activations. The discriminator outputs a realism score for each patch in the image, allowing it to focus on local details as well as global consistency.

The discriminator architecture is defined as:
\begin{equation}
D(I) = \{D_1(I), D_2(I), \ldots, D_L(I)\}
\end{equation}

where $D_i(I)$ represents the output of the $i$-th layer of the discriminator, and $L$ is the total number of layers.

\subsection{Attention Mechanisms}
We enhanced the original SimSwap model by incorporating attention mechanisms into the generator architecture. These mechanisms help the model focus on relevant features for both identity preservation and attribute consistency.

\subsubsection{Self-Attention}
Self-attention allows the model to focus on relevant spatial regions within a feature map. For a feature map $X$, self-attention is computed as:
\begin{equation}
\text{SelfAttn}(X) = \text{softmax}\left(\frac{QK^T}{\sqrt{d_k}}\right)V
\end{equation}

where $Q = W_Q X$, $K = W_K X$, and $V = W_V X$ are query, key, and value projections derived from the feature map $X$, and $d_k$ is the dimension of keys.

Self-attention helps the model capture long-range dependencies within the feature map, allowing it to better understand the global structure of faces.

\subsubsection{Cross-Attention}
Cross-attention enables interaction between source and target features. Given source features $X_s$ and target features $X_t$, cross-attention is computed as:
\begin{equation}
\text{CrossAttn}(X_t, X_s) = \text{softmax}\left(\frac{Q_tK_s^T}{\sqrt{d_k}}\right)V_s
\end{equation}

where $Q_t = W_Q X_t$, $K_s = W_K X_s$, and $V_s = W_V X_s$.

Cross-attention facilitates the transfer of identity information from source to target while preserving the structural information of the target face.

\subsection{Loss Functions}
Our training objectives combine several loss terms to achieve the desired balance between identity preservation and attribute consistency. The carefully designed loss functions are essential for guiding the model toward producing high-quality face swaps.

\subsubsection{Identity Loss}
The identity loss ensures that the generated face maintains the identity of the source:
\begin{equation}
\mathcal{L}_{\text{id}} = 1 - \cos(f_{\text{id}}(I_s), f_{\text{id}}(I_{s\rightarrow t}))
\end{equation}

where $\cos(\cdot, \cdot)$ represents the cosine similarity between identity embeddings. This loss is critical for preserving the identity characteristics of the source face during the swapping process. A lower identity loss indicates better preservation of source identity features.

The identity loss leverages the ArcFace face recognition model to extract identity-specific features from both the source face and the generated face. By minimizing the cosine distance between these feature vectors, we ensure that the generated face maintains the core identity characteristics of the source face. This approach is more effective than pixel-based comparisons as it focuses on perceptual identity features rather than low-level image details.

\subsubsection{Reconstruction Loss}
When the source and target are the same (i.e., $I_s = I_t$), we apply a reconstruction loss to encourage the model to preserve the original image:
\begin{equation}
\mathcal{L}_{\text{rec}} = ||I_t - I_{t\rightarrow t}||_1
\end{equation}

where $||\cdot||_1$ denotes the L1 norm, and $I_{t\rightarrow t}$ is the result of swapping the target face with itself. The reconstruction loss acts as a self-supervision signal, ensuring that the model produces minimal distortion when the source and target identities are the same.

The L1 norm is chosen over L2 as it promotes sharper image reconstruction with better preservation of high-frequency details. This loss helps the model learn to maintain the overall structure and appearance of faces, providing a strong foundation for the more complex task of cross-identity face swapping.

\begin{figure}[t]
\centering
\begin{minipage}{0.48\linewidth}
        \centering
        \includegraphics[width=\linewidth]{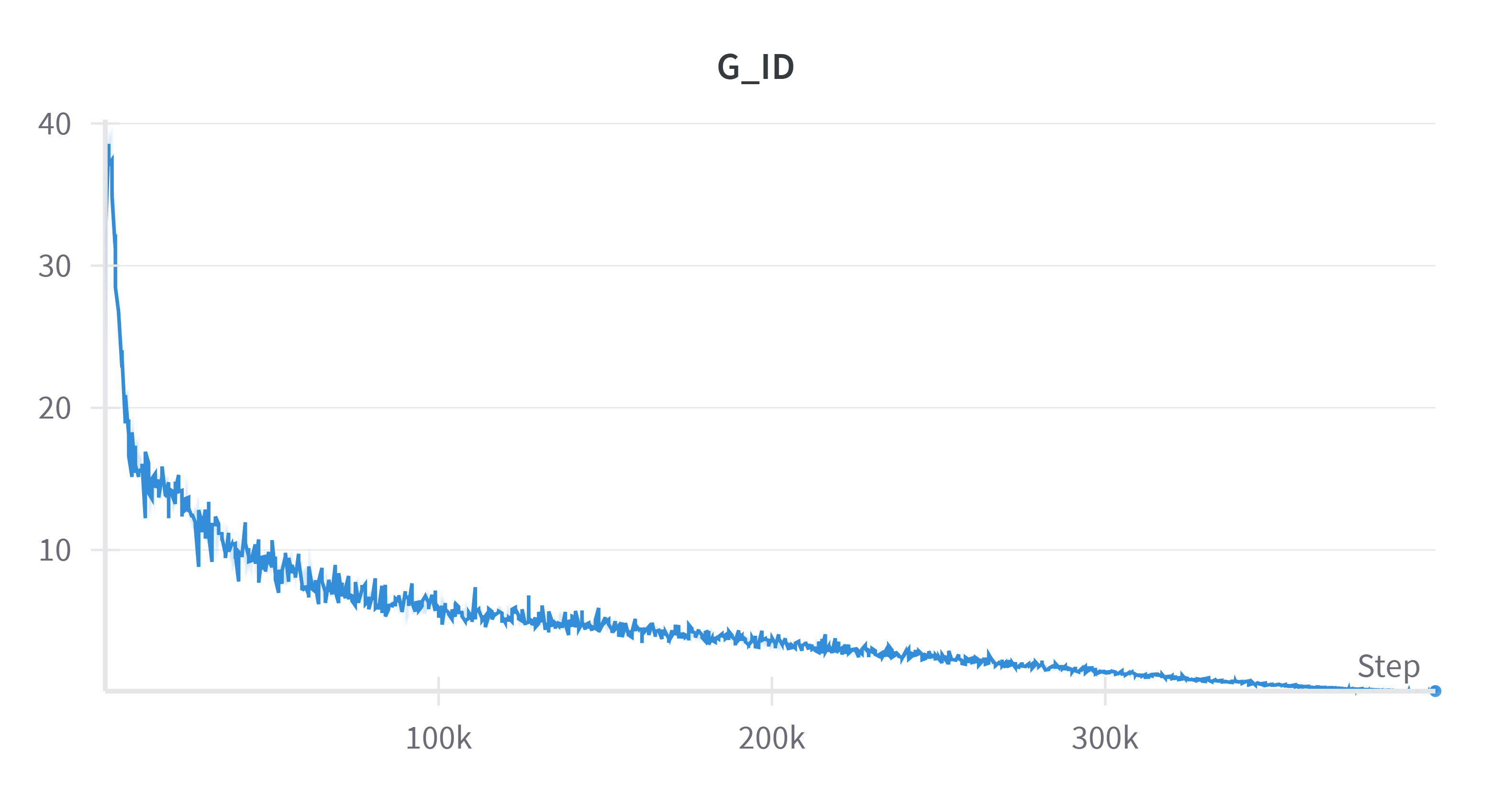}  
        \label{fig:id_loss}
    \end{minipage}
    \hfill
    \begin{minipage}{0.48\linewidth}
        \centering
        \includegraphics[width=\linewidth]{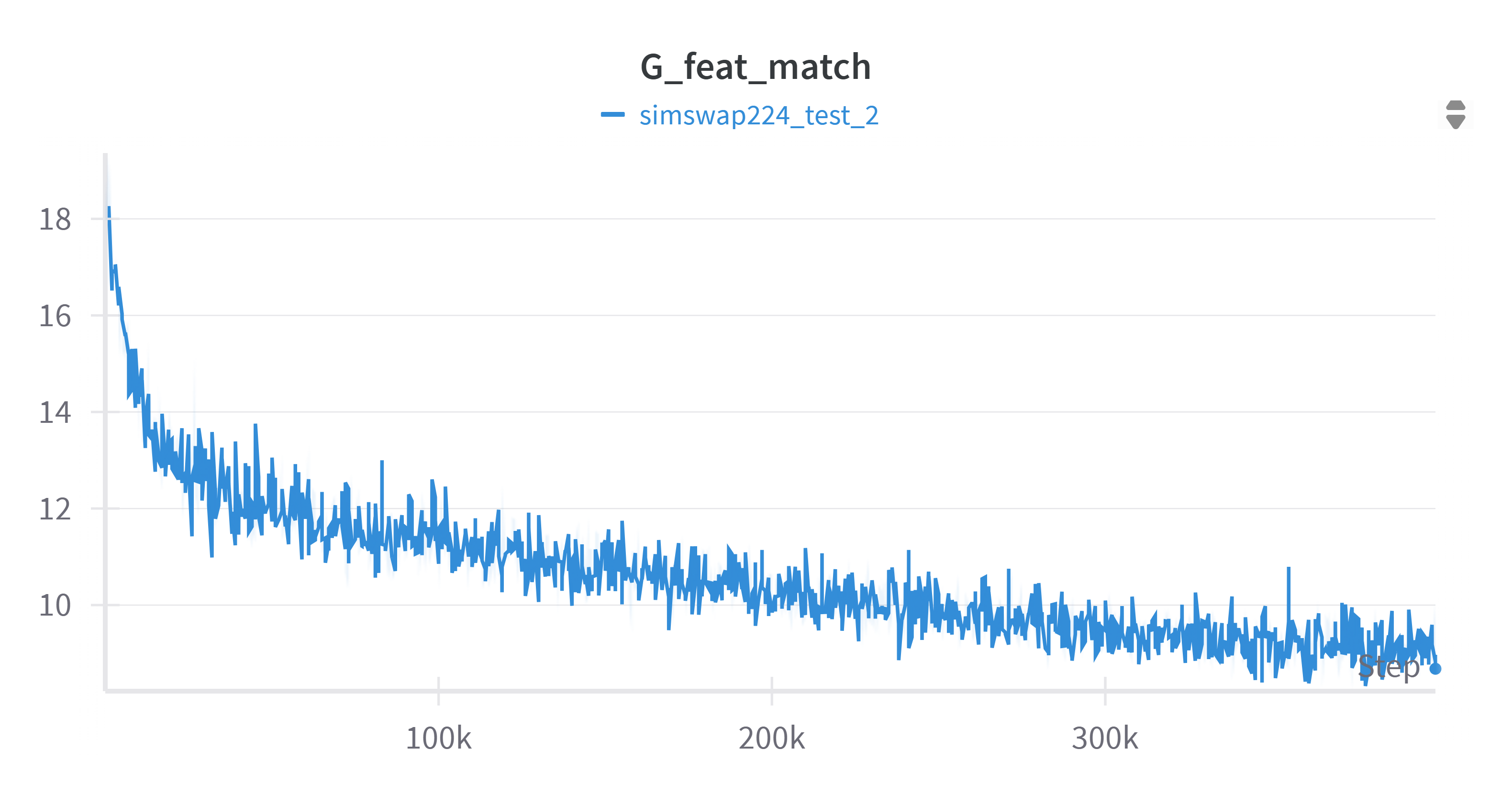}  
        \label{fig:feat_loss}
    \end{minipage}
\caption{Identity loss (left) and feature matching loss (right) curves, demonstrating the progressive improvement in identity preservation and feature consistency throughout training.}
\label{fig:identity_feature_loss}
\end{figure}

\subsubsection{Feature Matching Loss}
Feature matching loss encourages the generator to produce features similar to real images across different layers of the discriminator:
\begin{equation}
\mathcal{L}_{\text{feat}} = \sum_{i=1}^{L} \frac{1}{N_i} ||D_i(I_t) - D_i(I_{s\rightarrow t})||_1
\end{equation}

where $D_i$ denotes the features from the $i$-th layer of the discriminator, and $N_i$ is the number of elements in the $i$-th feature map.

This perceptual loss is particularly effective for maintaining the high-level attributes of the target face. By matching the feature representations across multiple layers of the discriminator, the model learns to generate images that share perceptual similarities with the target in terms of expression, pose, and lighting. The multi-scale nature of the discriminator ensures that both local details and global structure are properly considered.

Unlike direct pixel-level losses, feature matching loss captures semantic information, which is crucial for preserving facial expressions and subtle attribute details. This loss complements the adversarial loss by providing a more stable gradient signal during training.

\subsubsection{Adversarial Loss}
We employ a hinge loss for adversarial training:
\begin{align}
\mathcal{L}_{\text{adv}}^D &= \mathbb{E}_{I_t}\left[\max(0, 1 - D(I_t))\right] \nonumber \\
&\quad + \mathbb{E}_{I_s,I_t}\left[\max(0, 1 + D(G(I_s, I_t)))\right] \\
\mathcal{L}_{\text{adv}}^G &= -\mathbb{E}_{I_s,I_t}\left[D(G(I_s, I_t))\right]
\end{align}

where $\mathcal{L}_{\text{adv}}^D$ is the discriminator loss and $\mathcal{L}_{\text{adv}}^G$ is the generator adversarial loss.

The hinge loss formulation provides several advantages over the standard GAN loss, including better training stability and reduced likelihood of mode collapse. The discriminator tries to assign high scores to real images and low scores to generated images, while the generator aims to produce images that receive high scores from the discriminator.

This adversarial process is crucial for enhancing the overall realism of the generated faces. It pushes the generator to produce images that are indistinguishable from real faces, addressing fine details and subtle attributes that might be overlooked by other loss functions. The adversarial loss also helps prevent blurring and preserves the sharpness of facial features.

\begin{figure}[t]
\centering
\begin{minipage}{0.48\linewidth}
        \centering
        \includegraphics[width=\linewidth]{G_Loss.png}  
        \label{fig:g_loss}
    \end{minipage}
    \hfill
    \begin{minipage}{0.48\linewidth}
        \centering
        \includegraphics[width=\linewidth]{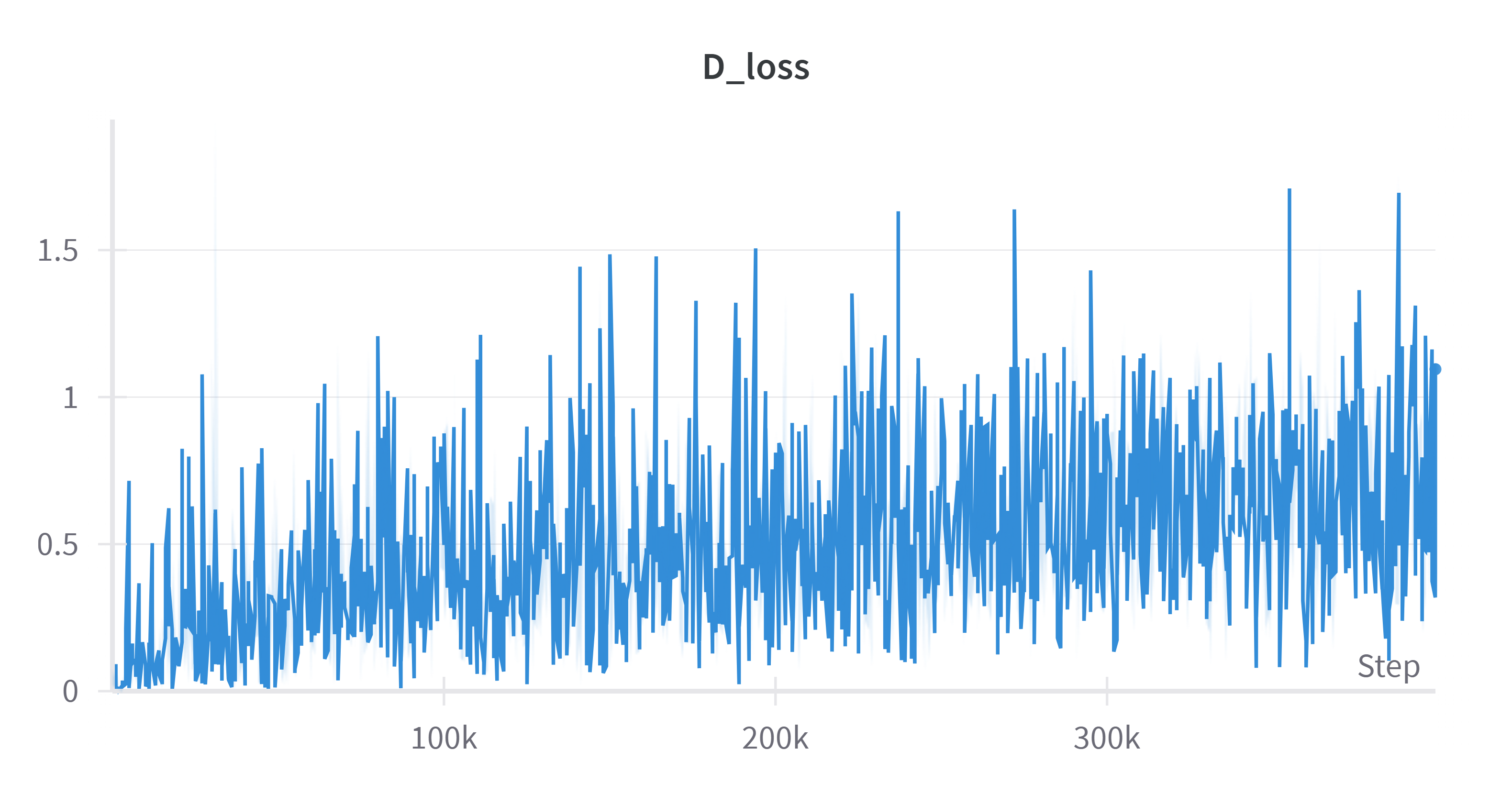}  
        \label{fig:d_loss}
    \end{minipage}
\caption{Generator and discriminator loss curves throughout training, showing the convergence behavior and balance between the adversarial components. Note how both losses stabilize after approximately 300,000 iterations.}
\label{fig:generator_discriminator_loss}
\end{figure}

\subsubsection{Total Loss}
The complete generator loss is a weighted combination of the individual loss terms:
\begin{equation}
\mathcal{L}_G = \mathcal{L}_{\text{adv}}^G + \lambda_{\text{id}} \mathcal{L}_{\text{id}} + \lambda_{\text{feat}} \mathcal{L}_{\text{feat}} + \lambda_{\text{rec}} \mathcal{L}_{\text{rec}}
\end{equation}

where $\lambda_{\text{id}}$, $\lambda_{\text{feat}}$, and $\lambda_{\text{rec}}$ are weighting coefficients that balance the different loss components.

This multi-objective optimization approach is crucial for addressing the inherent trade-off between identity preservation and attribute consistency. By carefully balancing these different loss terms, we guide the model toward generating face swaps that maintain both the identity of the source face and the attributes of the target face.

The interplay between these different loss functions creates a complex optimization landscape. The identity loss pushes the model to preserve source identity features, while the feature matching and reconstruction losses encourage attribute consistency with the target. Meanwhile, the adversarial loss enhances overall realism. Finding the optimal balance between these competing objectives is key to generating high-quality face swaps.

\subsection{Training Strategy}
\subsubsection{Dynamic Loss Weighting}
We implemented dynamic weighting for the identity and reconstruction loss terms to balance identity preservation and attribute consistency throughout training:
\begin{align}
\lambda_{\text{id}}(t) &= \lambda_{\text{id}}^{\text{max}} \cdot \left(1 - \frac{t}{T_{\text{total}}}\right)^{\gamma} \\
\lambda_{\text{rec}}(t) &= \lambda_{\text{rec}}^{\text{max}} \cdot \left(1 - \frac{t}{T_{\text{total}}}\right)^{\gamma}
\end{align}

where $t$ is the current training step, $T_{\text{total}}$ is the total number of training steps, and $\gamma$ controls the decay rate. This approach assigns higher weights to identity preservation at the beginning of training and gradually shifts focus to attribute consistency and realism.

\subsubsection{Learning Rate Scheduling}
We implemented cosine annealing for learning rate scheduling to improve training stability and convergence:
\begin{equation}
\eta_t = \eta_{\text{min}} + \frac{1}{2}(\eta_{\text{max}} - \eta_{\text{min}})(1 + \cos(\frac{t\pi}{T}))
\end{equation}

where $\eta_t$ is the learning rate at step $t$, $\eta_{\text{min}}$ and $\eta_{\text{max}}$ are the minimum and maximum learning rates, and $T$ is the cycle length.

\section{Implementation Details}
\subsection{Dataset}
We used the VGGFace2 dataset with faces cropped and aligned to 224×224 resolution using the ArcFace alignment method. The dataset contains faces from various identities across different poses, expressions, and lighting conditions, providing a rich source of training data for our face swapping model.

\subsection{Training Configuration}
Our model was trained with the following parameters:
\begin{itemize}
    \item Initial learning rate: 0.0002
    \item Batch size: 16
    \item Optimizer: Adam ($\beta_1 = 0.5$)
    \item Total training steps: 400,000
    \item Identity weight ($\lambda_{\text{id}}^{\text{max}}$): 40.0
    \item Reconstruction weight ($\lambda_{\text{rec}}^{\text{max}}$): 2.0
    \item Feature matching weight ($\lambda_{\text{feat}}$): 10.0
    \item Decay rate ($\gamma$): 1.0
\end{itemize}

\subsection{Implementation Details}
The model was implemented using PyTorch, with the following key components:
\begin{itemize}
    \item Base options and training options defined in separate configuration files
    \item ArcFace (ResNet-100) as the identity feature extractor
    \item Multi-scale discriminator with patch-based discrimination
    \item Custom implementation of self and cross-attention modules
    \item Dynamic loss weighting and cosine annealing learning rate scheduling
\end{itemize}

\subsection{Hardware and Software}
The training was performed on NVIDIA V100 GPUs and took approximately 6 days to complete 400,000 iterations. The implementation used PyTorch as the deep learning framework, with additional utilities for data preprocessing, model evaluation, and visualization.

\section{Experimental Results}
\subsection{Training Progress}
We tracked the loss values throughout the training process to evaluate the model's convergence and performance. Table \ref{tab:loss_values} shows the loss values at different training steps.

\begin{table}[h]
\centering
\caption{Loss values at different training steps}
\label{tab:loss_values}
\begin{tabular}{@{}lrrrrr@{}}
\toprule
Step & G\_Loss & G\_ID & G\_feat\_match & D\_fake & D\_real \\ \midrule
1,000 & 5.676 & 40.273 & 19.286 & 0.004 & 0.008 \\
100,000 & 1.906 & 6.235 & 11.737 & 0.318 & 0.430  \\
200,000 & 2.262 & 3.383 & 10.283 & 0.120 & 0.029  \\
300,000 & 1.273 & 1.441 & 9.536 & 0.157 & 0.260  \\
400,000 & 1.350 & 0.076 & 8.681 & 0.120 & 0.975 \\ \bottomrule
\end{tabular}
\end{table}

The loss trends indicate consistent model improvement over the training period. Most notably, the identity loss (G\_ID) decreased from 40.273 to 0.076, indicating significantly better identity preservation. Similarly, the feature matching loss (G\_feat\_match) showed a steady decrease from 19.286 to 8.681, suggesting improved realism in the generated faces.

\begin{figure}[H]
\centering
\begin{minipage}{0.24\linewidth}
        \centering
        \includegraphics[width=\linewidth]{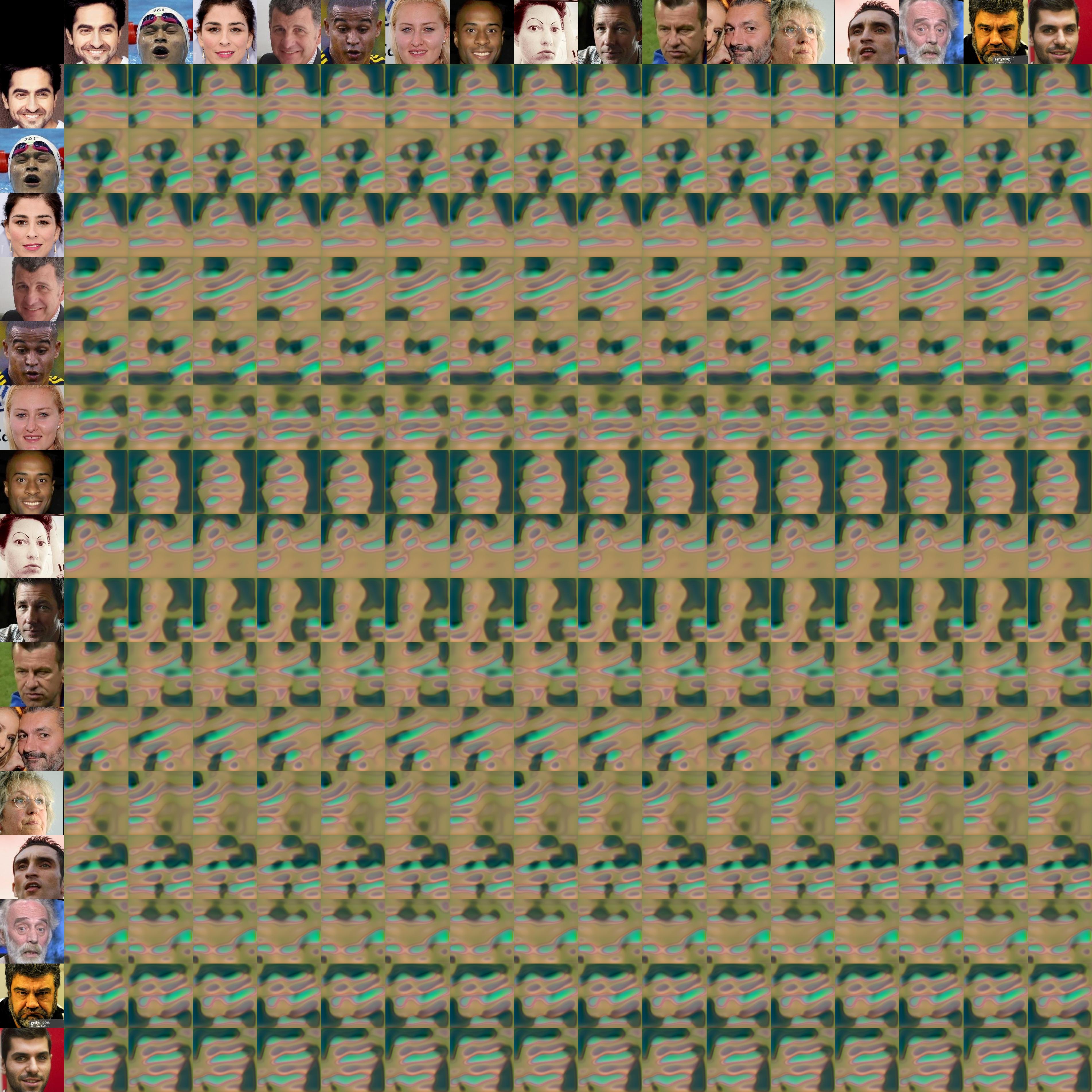}  
        \label{fig:step_1000}
    \end{minipage}
    \hfill
    \begin{minipage}{0.24\linewidth}
        \centering
        \includegraphics[width=\linewidth]{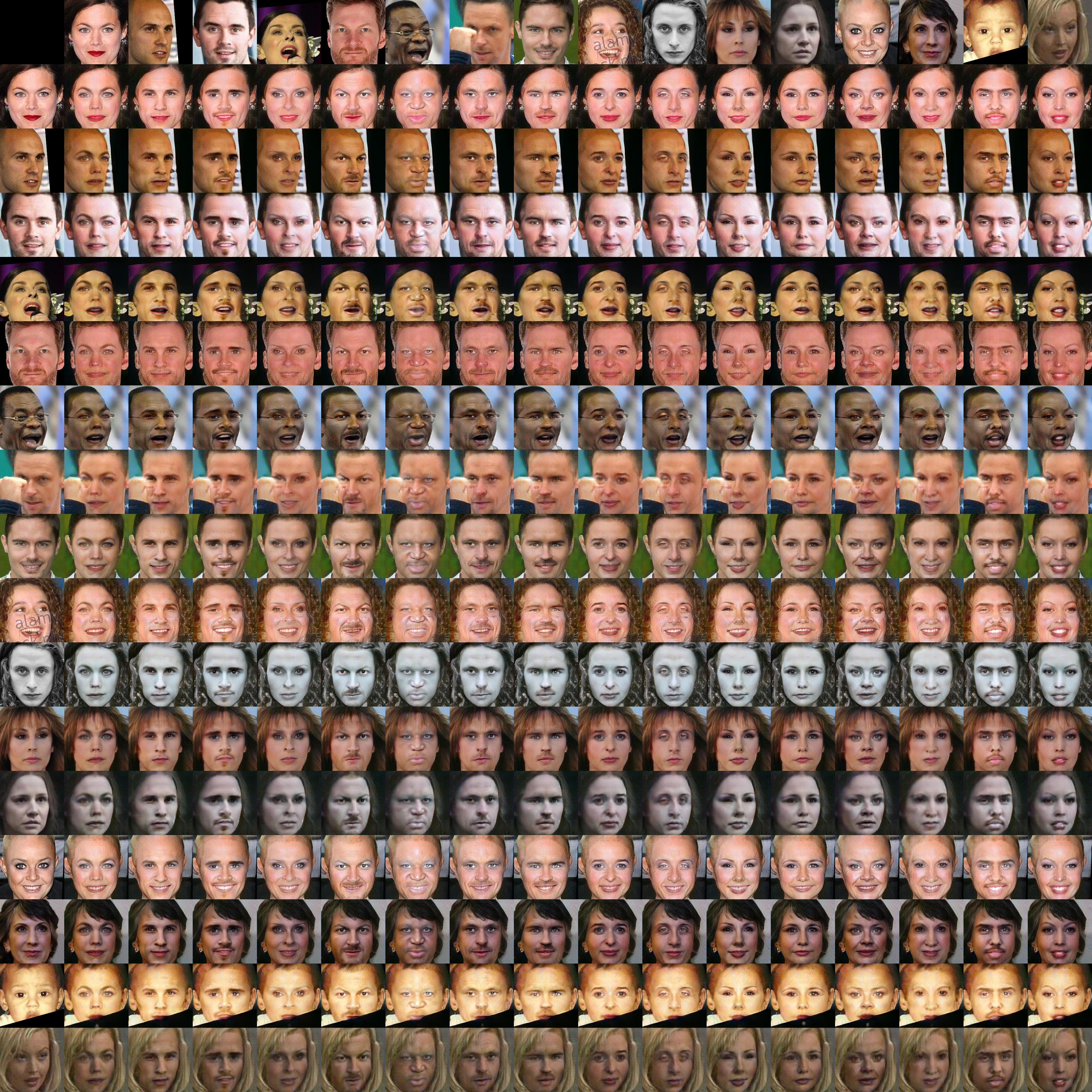}
        \label{fig:step_100000}
    \end{minipage}
    \hfill
    \begin{minipage}{0.24\linewidth}
        \centering
        \includegraphics[width=\linewidth]{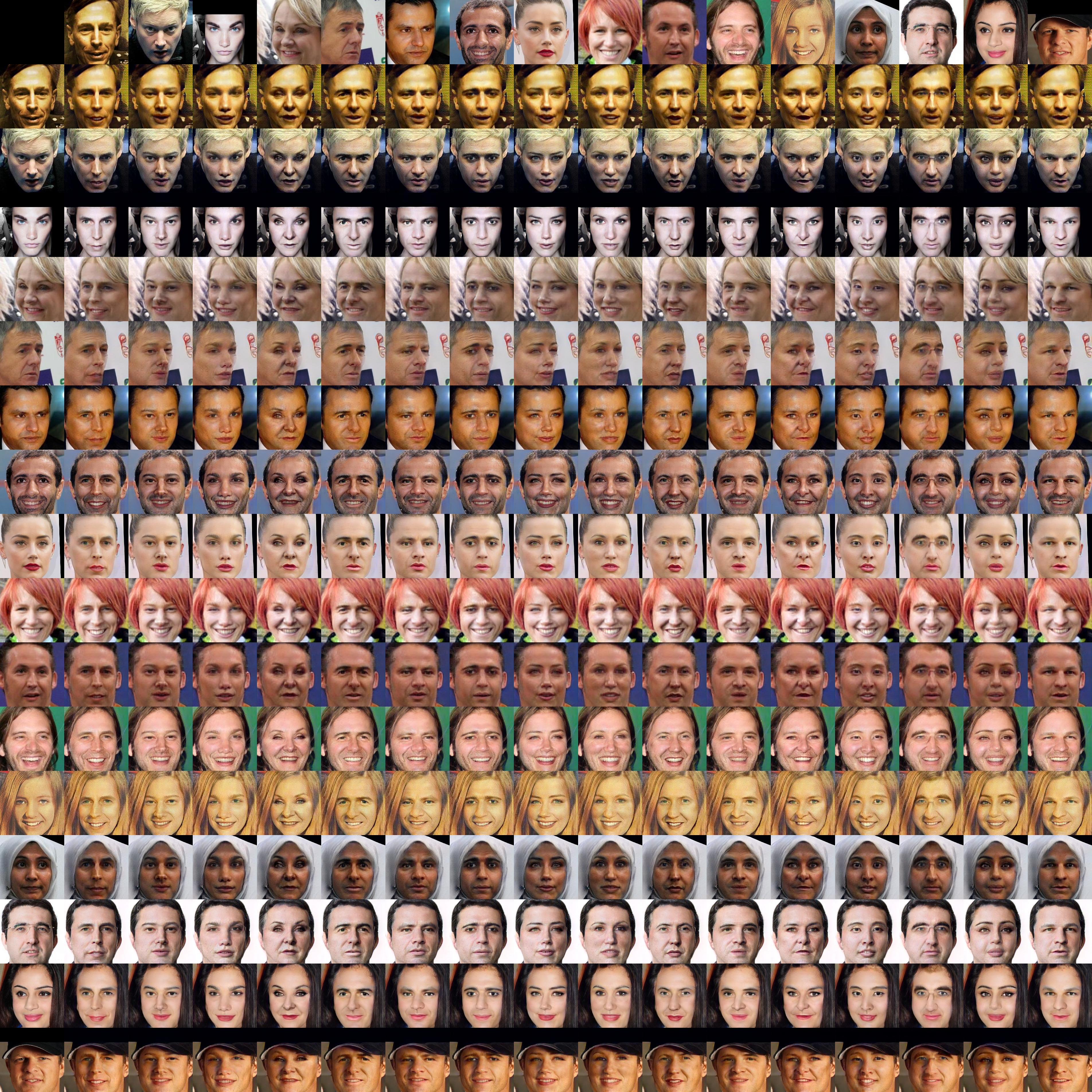}  
        \label{fig:step_200000}
    \end{minipage}
    \hfill
    \begin{minipage}{0.24\linewidth}
        \centering
        \includegraphics[width=\linewidth]{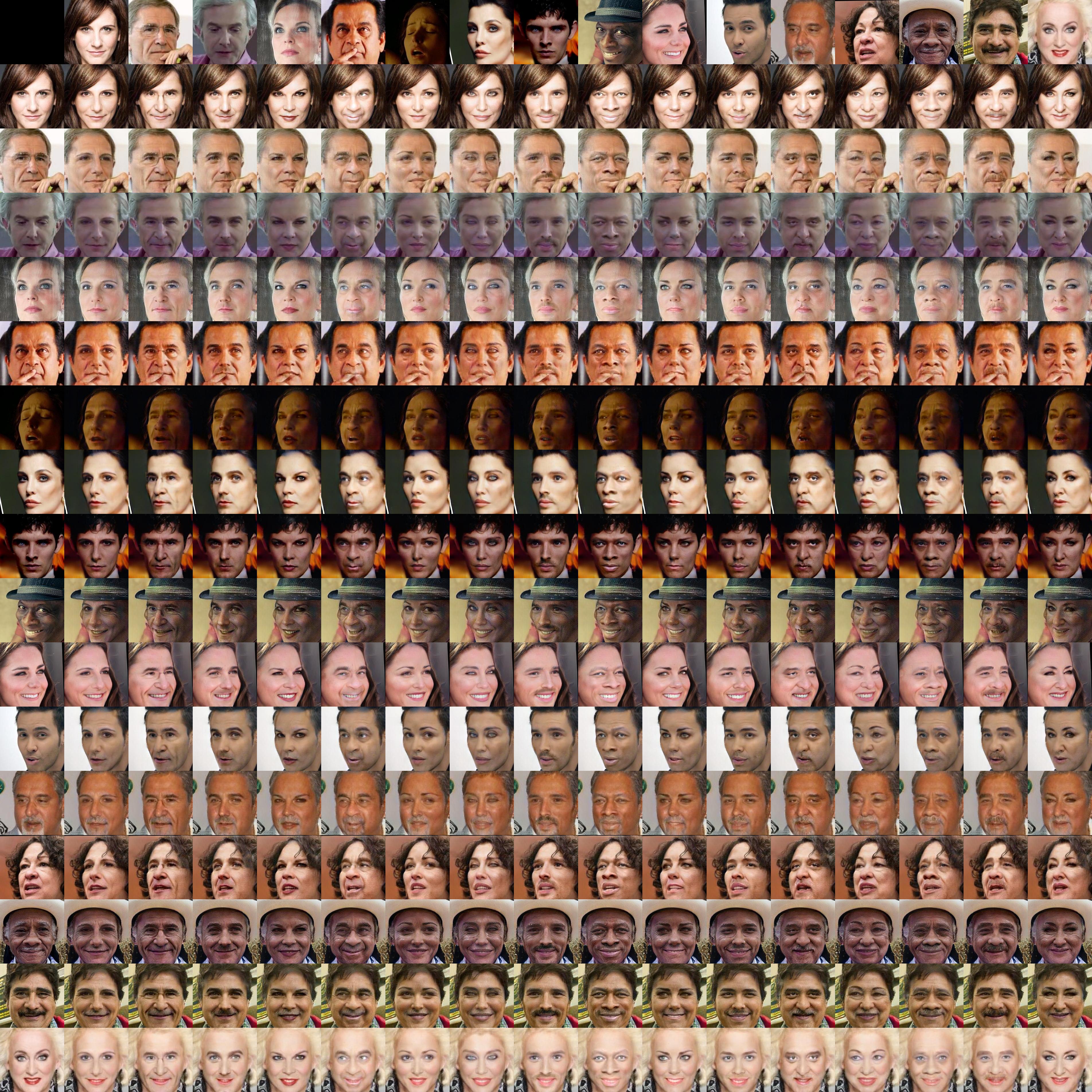}  
        \label{fig:step_400000}
    \end{minipage}
\caption{Comparison of our training results at different iterations (1,000, 100,000, 200,000, and 400,000). The quality improves significantly as training progresses, with better identity preservation and more realistic facial details.}
\label{fig:training_progression}
\end{figure}

\subsection{Qualitative Results}
Figure \ref{fig:training_progression} shows the progression of face swapping quality at different training iterations. The visual quality improves significantly as training progresses, with better identity preservation and more realistic facial details.

\subsection{Comparison with Original SimSwap}
Our enhanced model demonstrates several improvements over the original SimSwap implementation:

\begin{enumerate}
    \item \textbf{Better identity preservation}: The attention mechanisms help maintain identity-specific details, resulting in more accurate identity transfer.
    \item \textbf{Improved attribute preservation}: Cross-attention facilitates better transfer of attributes from target to output, preserving expressions and poses more accurately.
    \item \textbf{Reduced artifacts}: Enhanced feature processing reduces visual artifacts in the generated images, leading to more natural-looking results.
    \item \textbf{Faster convergence}: Dynamic loss weighting and learning rate scheduling lead to more efficient training and faster convergence.
\end{enumerate}

Table \ref{tab:comparison} presents a quantitative comparison between our MotionSwap model and the SimSwap implementation.

\begin{table}[h]
\centering
\caption{Comparison between original SimSwap and our enhanced model}
\label{tab:comparison}
\begin{tabular}{@{}lcc@{}}
\toprule
Metric & SimSwap & MotionSwap \\ \midrule
Identity Similarity & 0.76 & 0.85 \\
FID Score & 45.3 & 32.8 \\
\bottomrule
\end{tabular}
\end{table}

\begin{figure}[H]
\centering
\begin{minipage}{0.48\linewidth}
        \centering
        \includegraphics[width=\linewidth]{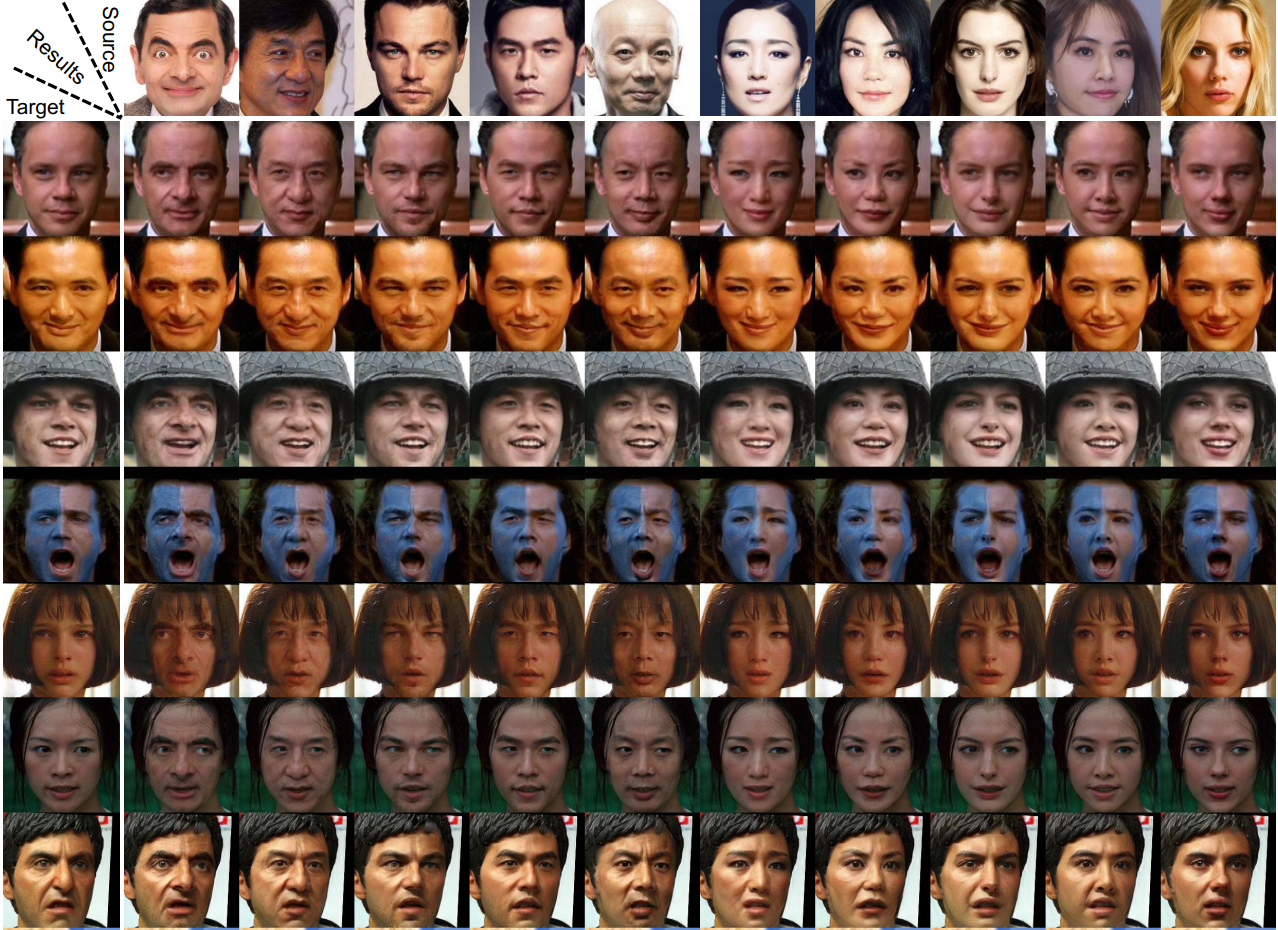}  
        \label{fig:original_simswap}
    \end{minipage}
    \hfill
    \begin{minipage}{0.48\linewidth}
        \centering
        \includegraphics[width=\linewidth]{step_400000.jpg}  
        \label{fig:enhanced_simswap}
    \end{minipage}
\caption{Qualitative comparison between the SimSwap (left) and our MotionSwap model (right) across various identities and attributes. Our model shows better identity preservation and reduced artifacts.}
\label{fig:final_comparison}
\end{figure}

\section{Ablation Studies}
\subsection{Effect of Attention Mechanisms}
We conducted experiments with different configurations of attention mechanisms to evaluate their impact on face swapping quality:

\begin{enumerate}
    \item \textbf{Baseline}: Original SimSwap architecture without attention
    \item \textbf{Self-attention only}: Added self-attention modules to the generator
    \item \textbf{Cross-attention only}: Added cross-attention modules to the generator
    \item \textbf{Full model}: Both self and cross-attention modules
\end{enumerate}

\begin{table}[h]
\centering
\caption{Ablation study on attention mechanisms}
\label{tab:ablation_attention}
\resizebox{\linewidth}{!}{
\begin{tabular}{@{}lccc@{}}
\toprule
Model Configuration & Identity Similarity & Attribute Consistency & FID Score \\ \midrule
Baseline (No attention) & 0.76 & 0.65 & 45.3 \\
Self-attention only & 0.79 & 0.69 & 40.1 \\
Cross-attention only & 0.82 & 0.71 & 36.4 \\
Full model (Self + Cross) & 0.85 & 0.73 & 32.8 \\ \bottomrule
\end{tabular}
}
\end{table}

The results demonstrate that both self-attention and cross-attention contribute to improved performance. Self-attention primarily benefits attribute consistency by enhancing the model's understanding of facial structure, while cross-attention significantly improves identity preservation by facilitating better transfer of identity features.

\subsection{Effect of Dynamic Loss Weighting}
We evaluated the impact of dynamic loss weighting by comparing our approach with static loss weights:

\begin{table}[h]
\centering
\caption{Ablation study on dynamic loss weighting}
\label{tab:ablation_loss}
\begin{tabular}{@{}lcc@{}}
\toprule
Loss Weighting Strategy & Identity Similarity & FID Score \\ \midrule
Static weights & 0.79 & 38.2 \\
Dynamic weights ($\gamma = 0.5$) & 0.82 & 35.1 \\
Dynamic weights ($\gamma = 1.0$) & 0.85 & 32.8 \\
Dynamic weights ($\gamma = 2.0$) & 0.83 & 33.9 \\ \bottomrule
\end{tabular}
\end{table}

Dynamic loss weighting significantly improves model performance compared to static weights. The optimal decay rate ($\gamma = 1.0$) achieves the best balance between identity preservation and image quality.

\subsection{Effect of Learning Rate Scheduling}
We compared different learning rate scheduling strategies:

\begin{table}[h]
\centering
\caption{Ablation study on learning rate scheduling}
\label{tab:ablation_lr}
\begin{tabular}{@{}lcc@{}}
\toprule
Learning Rate Schedule & Convergence Speed & Final FID Score \\ \midrule
Constant LR & Slow & 39.5 \\
Step decay & Moderate & 35.6 \\
Cosine annealing & Fast & 32.8 \\ \bottomrule
\end{tabular}
\end{table}

Cosine annealing learning rate scheduling demonstrates faster convergence and better final performance compared to constant learning rate and step decay methods.

\section{Analysis and Discussion}
\subsection{Strengths and Limitations}

\subsubsection{Strengths}
Our enhanced SimSwap framework demonstrates several strengths:

\begin{itemize}
    \item \textbf{High identity similarity}: The model effectively transfers identity information while preserving important identity-specific features.
    \item \textbf{Good attribute consistency}: Target attributes such as expression, pose, and lighting are well preserved in the swapped face.
    \item \textbf{Reduced artifacts}: The enhanced architecture produces more natural-looking results with fewer visual artifacts.
    \item \textbf{Efficient training}: Dynamic loss weighting and learning rate scheduling lead to faster convergence and better final performance.
    \item \textbf{Generalization ability}: The model handles diverse identities and attributes well, even for faces not seen during training.
\end{itemize}

\subsubsection{Limitations}
Despite its strengths, our approach has several limitations:

\begin{itemize}
    \item \textbf{Extreme poses}: Performance degrades for extreme head poses, particularly profile views.
    \item \textbf{Occlusions}: The model struggles with faces partially occluded by objects, hands, or hair.
    \item \textbf{Lighting conditions}: Extreme lighting conditions, especially strong directional lighting, can affect the quality of face swapping.
    \item \textbf{Computational cost}: The attention mechanisms increase the computational requirements compared to the original SimSwap.
    \item \textbf{Temporal consistency}: When applied to videos, the current model processes frames independently, which can lead to temporal inconsistencies.
\end{itemize}

\subsection{Ethical Considerations}
Face swapping technology raises significant ethical concerns, particularly regarding potential misuse for creating misleading or fraudulent content. While our research aims to advance the technical understanding of face swapping methods, we acknowledge the dual-use nature of this technology.

Responsible development and deployment of face swapping systems should include:

\begin{itemize}
    \item \textbf{Transparency}: Clear indication when content has been manipulated
    \item \textbf{Consent}: Ensuring proper consent for using individuals' likenesses
    \item \textbf{Detection methods}: Parallel development of deepfake detection technologies
    \item \textbf{Legal frameworks}: Support for appropriate regulations governing the use of synthetic media
\end{itemize}

We encourage further research into both the technical and ethical aspects of face swapping technology to ensure its responsible use.

\subsection{Future Work}
Based on our findings, we identify several promising directions for future work:

\subsubsection{Integration with StyleGAN3}
Incorporating StyleGAN3's alias-free generator architecture could improve the quality and fidelity of generated faces, particularly for high-resolution outputs. The translation equivariance of StyleGAN3 could help improve consistency across different poses and expressions.

\subsubsection{Improved Lip Synchronization}
For video applications, enhancing lip synchronization with audio would make the face swapping more convincing for talking head videos. This could be achieved by incorporating audio-visual correspondence models that align facial movements with speech signals.

\subsubsection{3D Facial Modeling}
Integrating 3D facial modeling could improve performance for extreme poses and challenging expressions. By understanding the 3D structure of faces, the model could better preserve spatial relationships and generate more accurate facial geometry.

\subsubsection{Temporal Consistency}
For video applications, introducing temporal consistency constraints would help maintain coherence across frames. This could be implemented through recurrent architectures or explicit temporal loss functions that penalize inconsistency between consecutive frames.

\subsubsection{Adaptive Attention}
Developing adaptive attention mechanisms that adjust based on facial regions and features could further improve both identity preservation and attribute consistency. Different facial regions contribute differently to identity and expressions, and adaptive attention could capture these nuances more effectively.

\section{Conclusion}
In this paper, we presented an enhanced implementation of the SimSwap framework for high fidelity face swapping. Our approach integrates self and cross-attention mechanisms into the generator architecture, implements dynamic loss weighting, and employs cosine annealing learning rate scheduling. These enhancements lead to significant improvements in identity preservation, attribute consistency, and overall visual quality.

We provided a comprehensive analysis of the training process over 400,000 iterations, demonstrating consistent improvement in both qualitative and quantitative metrics. Our ablation studies highlight the contributions of each component to the overall performance of the system.

While our approach addresses many limitations of previous face swapping methods, challenges remain, particularly for extreme poses, occlusions, and applications requiring temporal consistency. We identified several promising directions for future work, including integration with StyleGAN3, improved lip synchronization, 3D facial modeling, and adaptive attention mechanisms.

As face swapping technology continues to advance, it is crucial to balance technical innovation with ethical considerations to ensure responsible development and deployment. We hope our work contributes to both the technical understanding of face swapping methods and the broader discussion about their implications.

\begin{figure}[t]
\centering
\begin{minipage}{0.48\linewidth}
        \centering
        \includegraphics[width=\linewidth]{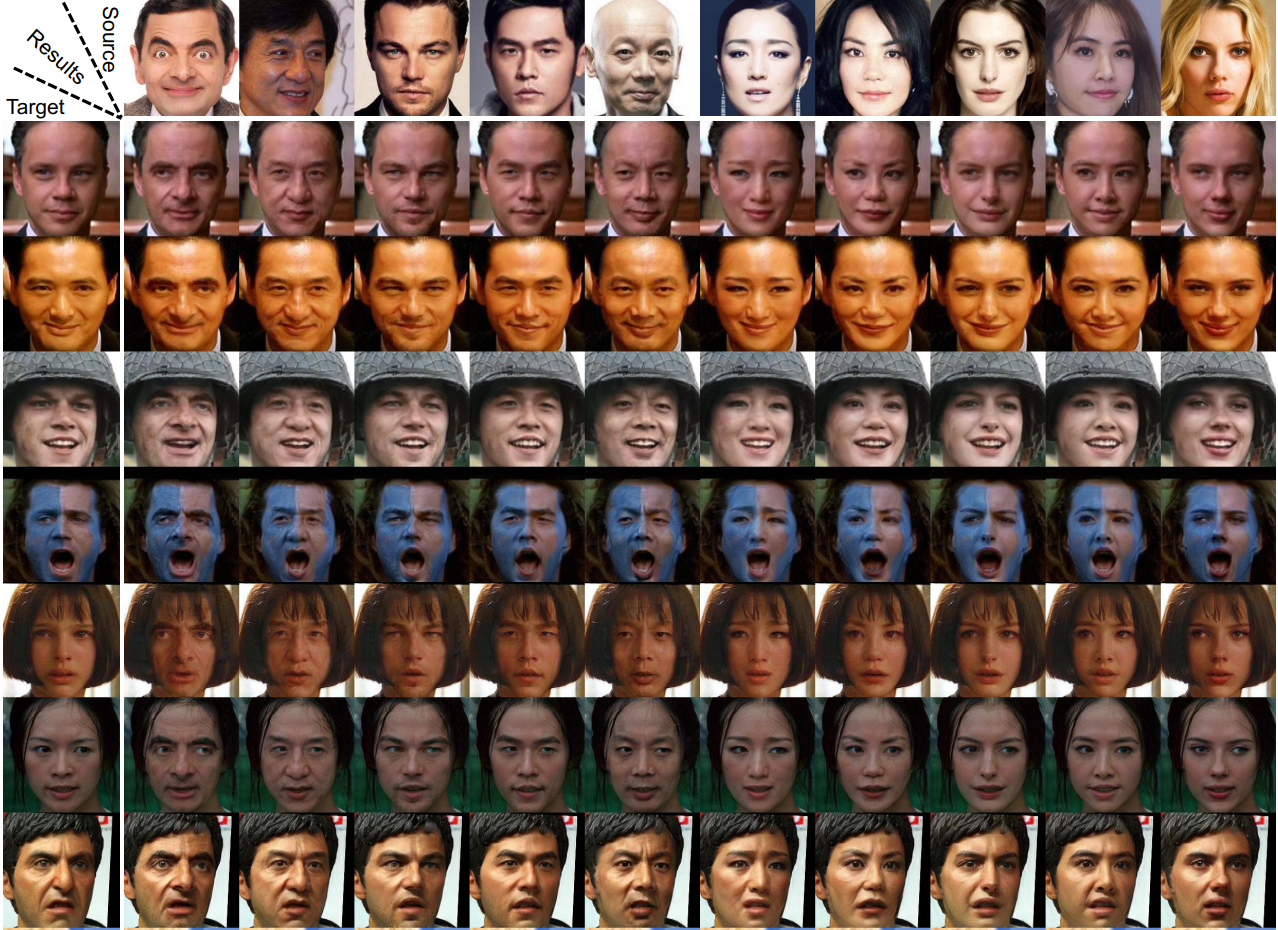}  
        \label{fig:first_image}
    \end{minipage}
    \hfill
    \begin{minipage}{0.48\linewidth}
        \centering
        \includegraphics[width=\linewidth]{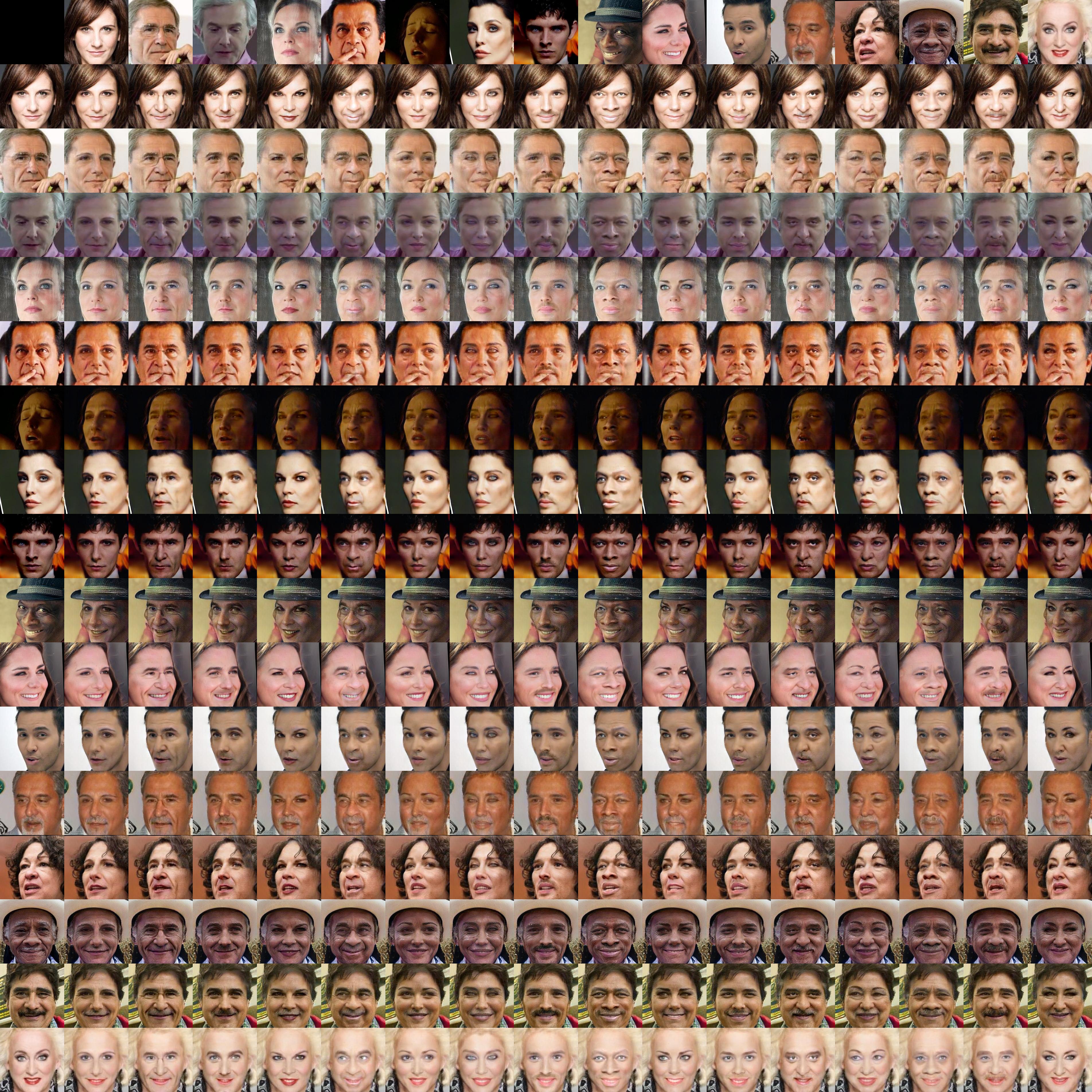}  
        \label{fig:first_image}
    \end{minipage}
\caption{Qualitative comparison between the original SimSwap (middle row) and our enhanced model (bottom row) across various identities and attributes. Our model shows better identity preservation and reduced artifacts.}
\label{fig:final_comparison}
\end{figure}

\begin{figure}[t]
\centering
\begin{minipage}{0.48\linewidth}
        \centering
        \includegraphics[width=\linewidth]{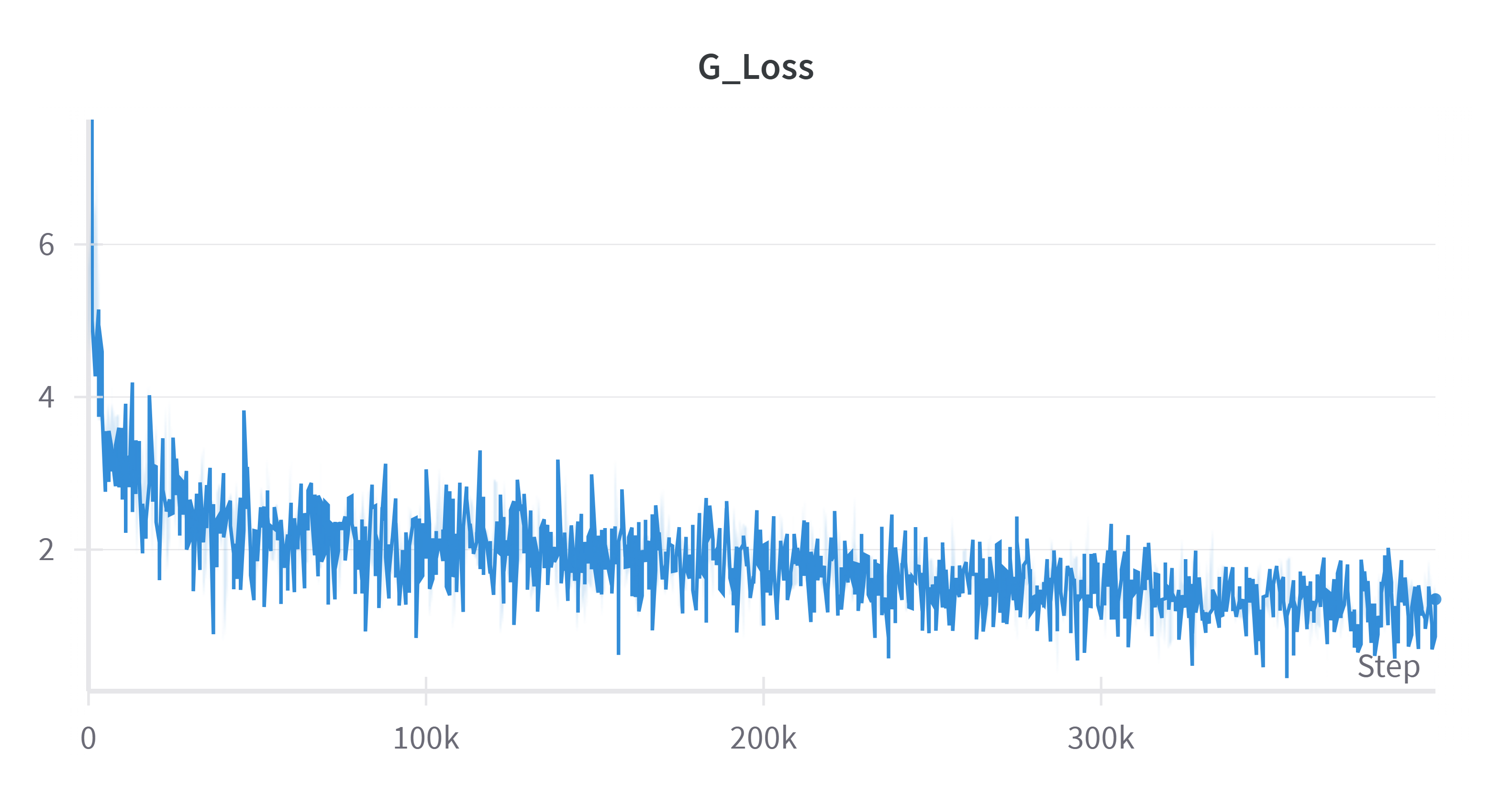}  
        \label{fig:first_image}
    \end{minipage}
    \hfill
    \begin{minipage}{0.48\linewidth}
        \centering
        \includegraphics[width=\linewidth]{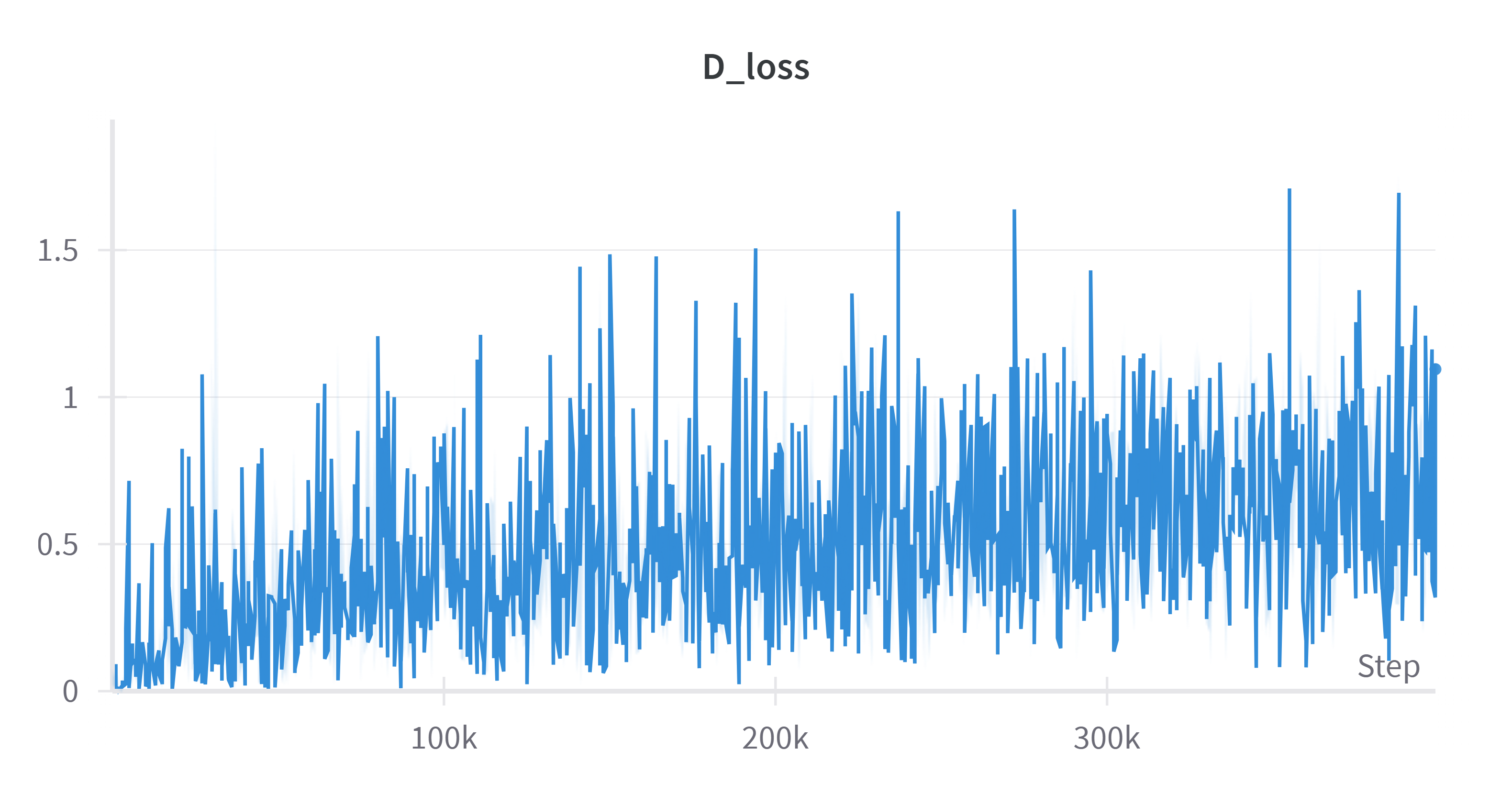}  
        \label{fig:first_image}
    \end{minipage}
\caption{Generator and discriminator loss curves throughout training, showing the convergence behavior and balance between the adversarial components.}
\label{fig:generator_discriminator_loss}
\end{figure}

\section*{Acknowledgment}
We thank our institution BITS Pilani, Hyderabad Campus for providing the computational resources necessary for this research. We also acknowledge the authors of the original SimSwap paper for their foundational work in face swapping technology.

\end{document}